\documentclass[11pt]{article}
\usepackage{acl2015}
\usepackage{times}
\usepackage{url}
\usepackage{latexsym}
\usepackage{enumitem}
\setlist{nolistsep}

\usepackage{breakcites}
\usepackage{bm}
\usepackage{amsfonts}
\usepackage{amsmath}
\usepackage{color}

\usepackage{graphicx} % more modern
\usepackage{epsfig} % less modern
\usepackage{subfigure}

\usepackage{epstopdf}
\usepackage{pdfpages}

\usepackage{array,booktabs}

\usepackage{algorithm}
\usepackage{algorithmic}
\usepackage{setspace}

\usepackage{todonotes}

\makeatletter
\newcommand{\@BIBLABEL}{\@emptybiblabel}
\newcommand{\@emptybiblabel}[1]{}
\makeatother

\usepackage{hyperref}

\usepackage[all]{hypcap}

\definecolor{Black}{rgb}{0,0,0}
\hypersetup{
  breaklinks=true,
  plainpages=false,
  colorlinks=true,
  anchorcolor=Black,
  linkcolor=Black,
  citecolor=Black,
  pdftitle={Lifelong Learning for Sentiment Classification},
  pdfauthor={Zhiyuan Chen, Nianzu Ma, Bing Liu},
  pdfsubject={Proceedings of ACL 2015},
  pdfkeywords={Lifelong Learning, Lifelong Machine Learning, Sentiment Classification}
}

\DeclareMathAlphabet{\mathbbmsl}{U}{bbm}{m}{sl}

\title{Lifelong Learning for Sentiment Classification}

\author{Zhiyuan Chen,\quad Nianzu Ma,\quad Bing Liu \\
	Department of Computer Science \\
	University of Illinois at Chicago \\
	{\tt \{czyuanacm,jingyima005\}@gmail.com,liub@cs.uic.edu} \\}

\date{}

\begin{document}
\maketitle
\begin{abstract}
This paper proposes a novel lifelong learning (LL) approach to sentiment classification. LL mimics the human continuous learning process, i.e., retaining the knowledge learned from past tasks and use it to help future learning. In this paper, we first discuss LL in general and then LL for sentiment classification in particular. The proposed LL approach adopts a Bayesian optimization framework based on stochastic gradient descent. Our experimental results show that the proposed method outperforms baseline methods significantly, which demonstrates that lifelong learning is a promising research direction.

\end{abstract}

\section{Introduction}
\label{sec:intro}

Sentiment classification is the task of classifying an opinion document as expressing a positive or negative sentiment.~\newcite{liu2012} and~\newcite{Pang2008Sentiment} provided good surveys of the existing research. In this paper, we tackle sentiment classification from a novel angle, \textit{lifelong learning} (LL), or \textit{lifelong machine learning}. This learning paradigm aims to learn as humans do: retaining the learned knowledge from the past and use the knowledge to help future learning~\cite{Thrun_1998LearningToLearn,Chen2014ICML,Silver2013}. 

Although many machine learning topics and techniques are related to LL, e.g., lifelong learning~\cite{Thrun_1998LearningToLearn,Chen2014ICML,Silver2013}, transfer learning~\cite{Jiang2008literature,Pan2010TLSurvey}, multi-task learning~\cite{Caruana1997}, never-ending learning~\cite{Carlson2010}, self-taught learning~\cite{Raina2007}, and online learning~\cite{Bottou1998},
there is still no unified definition for LL. 

Based on the prior work and our research, to build an LL system, we believe that we need to answer the following key questions:

\begin{enumerate}[leftmargin=*]
	
\item What information should be retained from the past learning tasks?

\item What forms of knowledge will be used to help future learning?

\item How does the system obtain the knowledge?

\item How does the system use the knowledge to help future learning?

\end{enumerate}

Motivated by these questions, we present the following definition of \textit{lifelong learning} (LL). 

\vspace{3pt}
\noindent
\textbf{Definition (Lifelong Learning)}: A learner has performed learning on a sequence of tasks, from 1 to $N-1$. When faced with the $N$th task, it uses the knowledge gained in the past $N-1$ tasks to help learning for the $N$th task. An LL system thus needs the following four general components:  

\begin{enumerate}[leftmargin=*]
	
\item \textit{Past Information Store} (\textit{PIS}): It stores the information resulted from the past learning. This may involve sub-stores for information such as (1) the original data used in each past task, (2) intermediate results from the learning of each past task, and (3) the final model or patterns learned from the past task, respectively. 

\item \textit{Knowledge Base} (\textit{KB}): It stores the knowledge mined or consolidated from PIS (Past Information Store). This requires a knowledge representation scheme suitable for the application. 

\item \textit{Knowledge Miner} (\textit{KM}). It mines knowledge from PIS (Past Information Store). This mining can be regarded as a meta-learning process because it learns knowledge from information resulted from learning of the past tasks. The knowledge is stored to KB (Knowledge Base).

\item \textit{Knowledge-Based Learner} (\textit{KBL}): Given the knowledge in KB, this learner is able to leverage the knowledge and/or some information in PIS for the new task.

\end{enumerate}
\vspace{3pt}

Based on this, we can define \textit{lifelong sentiment classification} (LSC):

\vspace{3pt}
\noindent
\textbf{Definition (Lifelong Sentiment Classification)}: A learner has performed a sequence of supervised sentiment classification tasks, from 1 to $N-1$, where each task consists of a set of training documents with positive and negative polarity labels. Given the $N$th task, it uses the knowledge gained in the past $N-1$ tasks to learn a better classifier for the $N$th task.
\vspace{3pt}

%state our specific problem as follows:
%\textbf{Problem Statement}. Assume that we have already performed supervised learning in many past domains. The knowledge learned in these past domain tasks is retained in a knowledge base called $KB$. $KB$ is then used to improve sentiment classification in the \textit{new} or \textit{target domain} $D^t$ which has labeled training data. After learning in domain $D^t$, its knowledge is also incorporated into $KB$.

It is useful to note that although many researchers have used transfer learning for supervised sentiment classification, LL is different from the classic transfer learning or domain adaptation~\cite{Pan2010TLSurvey}. Transfer learning typically uses labeled training data from one (or more) source domain(s) to help learning in the target domain that has little or no labeled data~\cite{Aue+Gamon:05a,Bollegala2011}. It does not use the results of the past learning or knowledge mined from the results of the past learning. Further, transfer learning is usually inferior to traditional supervised learning when the target domain already has good training data. In contrast, our target (or future) domain/task has good training data and we aim to further improve the learning using both the target domain training data and the knowledge gained in past learning. To be consistent with prior research, we treat the classification of one domain as one learning task.

One question is why the past learning tasks can contribute to the target domain classification given that the target domain already has labeled training data. The key reason is that the training data may not be fully representative of the test data due to the \textit{sample selection bias}~\cite{heckman1979sample,shimodaira2000improving,zadrozny2004learning}. In few real-life applications, the training data are fully representative of the test data. For example, in a sentiment classification application, the test data may contain some sentiment words that are absent in the training data of the target domain, while these sentiment words have appeared in some past domains. So the past domain knowledge can provide the prior polarity information in this situation.

Like most existing sentiment classification papers~\cite{liu2012}, this paper focuses on binary classification, i.e., positive ($+$) and negative ($-$) polarities. But the proposed method is also applicable to multi-class classification. To embed and use the knowledge in building the target domain classifier, we propose a novel optimization method based on the Na\"{i}ve Bayesian (NB) framework and stochastic gradient descent. The knowledge is incorporated using penalty terms in the optimization formulation. This paper makes three contributions:

\begin{enumerate}[leftmargin=*]

\item It proposes a novel lifelong learning approach to sentiment classification, called \textit{lifelong sentiment classification} (LSC). 
\item It proposes an optimization method that uses
penalty terms to embed the knowledge gained
in the past and to deal with domain dependent
sentiment words to build a better classifier.
\item It creates a large corpus containing reviews from 20 diverse product domains for extensive evaluation. The experimental results demonstrate the superiority of the proposed method.
\end{enumerate}

\section{Related Work}

Our work is mainly related to lifelong learning and multi-task learning~\cite{Thrun_1998LearningToLearn,Caruana1997,Chen2014ICML,Silver2013}. Existing lifelong learning approaches focused on exploiting invariances~\cite{Thrun_1998LearningToLearn} and other types of knowledge~\cite{Chen2014ICML,Chen2014KDD,Ruvolo2013ICML} across multiple tasks. Multi-task learning optimizes the learning of multiple related tasks at the same time~\cite{Caruana1997,Chen2011,saha2011online,zhang2008flexible}. However, these methods are not for sentiment analysis. Also, our na\"{i}ve Bayesian optimization based LL method is quite different from all these existing techniques.

Our work is also related to transfer learning or domain adaptation~\cite{Pan2010TLSurvey}. In the sentiment classification context,~\newcite{Aue+Gamon:05a} trained sentiment classifiers for the target domain using various mixes of labeled and unlabeled reviews.~\newcite{Blitzer2007} proposed to first find some common or pivot features from the source and the target, and then identify correlated features with the pivot features. The final classifier is built using the combined features.~\newcite{li2008multi} built a meta-classifier (called CLF) using the outputs of each base classifier constructed in each domain. Other works along similar lines include~\cite{Andreevskaia2008,Bollegala2011,He2011,Ku2009,Li2012,Li2013,Pan2010TLSurvey,Tan2007,Wu:2009:GRS:1667583.1667681,xia2011pos,yoshida2011transfer}. Additional details about these and other related works can be found in~\cite{liu2012}. However, as we discussed in the introduction, these methods do not focus on the ability to accumulate learned knowledge and leverage it in new learning in a lifelong manner.

\section{Proposed LSC Technique}
\subsection{Na\"{i}ve Bayesian Text Classification }

Before presenting the proposed method, we briefly review the Na\"{i}ve Bayesian (NB) text classification as our method uses it as the foundation.

NB text classification (McCallum and Nigam, 1998) basically computes the conditional probability of each word $w$ given each class $c_j$ (i.e., $P\left( {w|{c_j}} \right)$) and the prior probability of each class $c_j$ (i.e., $P(c_j)$), which are used to calculate the posterior probability of each class $c_j$ given a test document $d$ (i.e., $P({c_j}|d)$). $c_j$ is either positive ($+$) or negative ($-$) in our case.

The key parameter $P(w|{c_j})$ is computed as:

\begin{equation} \label{eq:1}
P\left( {w|{c_j}} \right) = \frac{{\lambda  + {N_{{c_j},w}}}}{{\lambda \left| V \right| + \mathop \sum \nolimits_{v = 1}^{|V|} {N_{{c_j},v}}}}
\end{equation}

\noindent where $N_{c_j,w}$ is the frequency of word $w$ in documents of class $c_j$. $|V|$ is the size of vocabulary $V$ and $\lambda$ ($0 \leq \lambda \leq 1$) is used for smoothing.

\subsection{Components in LSC}
\label{sec:componentsinlsc}

This subsection describes our proposed method corresponding to the proposed LL components.

\begin{enumerate}[leftmargin=*]
	
\item Past Information Store (PIS): In this work, we do not store the original data used in the past learning tasks, but only their results. For each past learning task $\hat{t}$, we store a) $P^{\hat{t}}(w|+)$ and $P^{\hat{t}}(w|-)$ for each word $w$ which are from task $\hat{t}$'s NB classifier (see Eq~\ref{eq:1}); and b) the number of times that $w$ appears in a positive ($+$) document $N^{\hat{t}}_{+,w}$ and the number of times that $w$ appears in a negative documents $N^{\hat{t}}_{-,w}$.

\item Knowledge Base (KB): Our knowledge base contains two types of knowledge:

\begin{enumerate}
	
	\item Document-level knowledge $N_{+,w}^{KB}$ (and $N_{-,w}^{KB})$: number of occurrences of $w$ in the documents of the positive (and negative) class in the past tasks, i.e., $N_{+,w}^{KB}=\sum_{\hat{t}} N^{\hat{t}}_{+,w}$ and $N_{-,w}^{KB}=\sum_{\hat{t}} N^{\hat{t}}_{-,w}$.
	
	\item Domain-level knowledge $M_{+,w}^{KB}$ (and $M_{-,w}^{KB})$: number of past tasks in which $P(w|+)>P(w|-)$ (and $P(w|+)<P(w|-)$).
	
\end{enumerate}

\item Knowledge Miner (KM). Knowledge miner is straightforward as it just performs counting and aggregation of information in PIS to generate knowledge (see 2(a) and 2(b) above).

\item Knowledge-Based Learner (KBL): This learner incorporates knowledge using regularization as penalty terms in our optimization. See the details in~\ref{sec:exploitingknowledge}.

\end{enumerate}
 
\subsection{Objective Function}
\label{sec:objectivefunction}

In this subsection, we introduce the objective function used in our method. The key parameters that affect NB classification results are $P(w|c_j )$ which are computed using empirical counts of word $w$ with class $c_j$, i.e., $N_{c_j,w}$ (Eq.~\ref{eq:1}). In binary classification, they are $N_{+,w}$ and $N_{-,w}$. This suggests that we can revise these counts appropriately to improve classification. In our optimization, we denote the optimized variables $X_{+,w}$ and $X_{-,w}$ as the number of times that a word $w$ appears in the positive and negative class. We called them \textit{virtual counts} to distinguish them from empirical counts $N_{+,w}$ and $N_{-,w}$. For correct classification, ideally, we should have the posterior probability $P({c_j}|{d_i})=1$ for labeled class $c_j$, and for the other class $c_f$, we should have $P({c_f}|{d_i})=0$. Formally, given a new domain training data $D^t$, our objective function is:

\begin{equation} \label{eq:2}
	\mathop \sum \limits_{i = 1}^{|{D^t}|} \left( {P\left( {{c_j}|{d_i}} \right) - P\left( {{c_f}|{d_i}} \right)} \right)
\end{equation}

Here $c_j$ is the actual labeled class of ${d_i} \in {D^t}$. In this paper, we use stochastic gradient descent (SGD) to optimize on the classification of each document ${d_i} \in {D^t}$. Due to the space limit, we only show the optimization process for a positive document (the process for a negative document is similar). The objective function under SGD for a positive document is:

\begin{equation} \label{eq:3}
	{F_{ + ,i}} = P\left( { + |{d_i}} \right) - P( - |{d_i})
\end{equation}

To further save space, we omit the derivation steps and give the final derivatives below (See the detailed derivation steps in the separate supplementary note):

\begin{equation} \label{eq:4}
\small
g\left( \boldsymbol{X} \right) = {\left( {\frac{{\lambda \left| V \right| + \mathop \sum \nolimits_{v = 1}^{|V|} {X_{ + ,v}}}}{{\lambda \left| V \right| + \mathop \sum \nolimits_{v = 1}^{|V|} {X_{ - ,v}}}}} \right)^{|{d_i}|}}
\end{equation}

\begin{equation} \label{eq:5}
\small
\begin{split}
\frac{\partial F_{+,i}} {\partial X_{+,u}} &=
\frac{ \frac{n_{u,d_i}}{\lambda+X_{+,u}} + \frac{P(-)}{P(+)} \prod_{w \in d_i} {\big( \frac{\lambda+X_{-,w}}{\lambda+X_{+,w}} \big)}^{n_{w,d_i}} \times \frac{\partial g} {\partial X_{+,u}} } {1+\frac{P(-)}{P(+)} \prod_{w \in d_i} {\big( \frac{\lambda+X_{-,w}}{\lambda+X_{+,w}} \big)}^{n_{w,d_i}} \times g(\bm X)} \\
&- \frac{n_{u,d_i}}{\lambda+X_{+,u}}
\end{split} 
\end{equation}

\begin{equation} \label{eq:6}
\small
\frac{\partial F_{+,i}} {\partial X_{-,u}}=
\frac{ \frac{n_{u,d_i}}{\lambda+X_{-,u}} \times g(\bm X) + \frac{\partial g}{\partial X_{-,u}} } {\frac{P(+)}{P(-)} \prod_{w \in d_i} {\big( \frac{\lambda+X_{+,w}}{\lambda+X_{-,w}} \big)}^{n_{w,d_i}} + g(\bm X)}
\end{equation}

\renewcommand{\arraystretch}{0.9} % Adjust the row spacing.
\renewcommand{\tabcolsep}{8pt} % Adjust the column spacing.
\begin{table*}[t]
	\small
	\begin{center}
		\begin{tabular}{|c|c|c|c|c|c|c|c|}
			\hline
			\rule{0pt}{10pt} Alarm Clock & 30.51 & Flashlight    & 11.69 & Home Theater System    & 28.84 & Projector   & 20.24 \\
			Baby        & 16.45 & GPS           & 19.50  & Jewelry                & 12.21 & Rice Cooker & 18.64 \\
			Bag         & 11.97 & Gloves        & 13.76 & Keyboard               & 22.66 & Sandal      & 12.11 \\
			Cable Modem & 12.53 & Graphics Card & 14.58 & Magazine Subscriptions & 26.88 & Vacuum      & 22.07 \\
			Dumbbell    & 16.04 & Headphone     & 20.99 & Movies TV              & 10.86 & Video Games & 20.93 \\ \hline
		\end{tabular}
	\end{center}
	\vspace{-2mm}
	\caption{\label{tab:20domains} Names of the 20 product domains and the proportion of negative reviews in each domain.}
	\vspace{-3mm}
\end{table*}

where $n_{u,d_i}$ is the term frequency of word $u$ in document $d_i$. $\bm X$ denotes all the variables consisting of $X_{+,w}$ and $X_{-,w}$ for each word $w$. The partial derivatives for a word $u$, i.e.,  $\frac{{\partial g}}{{\partial {X_{ + ,u}}}}$ and $\frac{{\partial g}}{{\partial {X_{ - ,u}}}}$, are quite straightforward and thus not shown here. $X_{+,w}^0=N_{+,w}^t+N_{+,w}^{KB}$ and $X_{-,w}^0=N_{-,w}^t+N_{-,w}^{KB}$ are served as a reasonable starting point for SGD, where $N_{+,w}^t$ and $N_{-,w}^t$ are the empirical counts of word $w$ and classes $+$ and $-$ from domain $D^t$, and $N_{+,w}^{KB}$ and $N_{-,w}^{KB}$ are from knowledge $KB$ (Section~\ref{sec:componentsinlsc}). The SGD runs iteratively using the following rules for the positive document $d_i$ until convergence, i.e., when the difference of Eq.~\ref{eq:2} for two consecutive iterations is less than $1e-3$ (same for the negative document), where $\gamma$ is the learning rate:

\begin{equation*}
\small
X_{+ ,u}^l = X_{ + ,u}^{l - 1} - \gamma \frac{{\partial {F_{ + ,i}}}}{{\partial {X_{ + ,u}}}}, X_{ - ,u}^l = X_{- ,u}^{l - 1} - \gamma \frac{{\partial {F_{ + ,i}}}}{{\partial {X_{ - ,u}}}}
\end{equation*}

\subsection{Exploiting Knowledge via Penalty Terms}
\label{sec:exploitingknowledge}

The above optimization is able to update the virtual counts for a better classification in the target domain. However, it does not deal with the issue of domain dependent sentiment words, i.e., some words may change the polarity across different domains. Nor does it utilize the domain-level knowledge in the knowledge base $KB$ (Section~\ref{sec:componentsinlsc}). We thus propose to add penalty terms into the optimization to accomplish these.

The intuition here is that if a word $w$ can distinguish classes very well from the target domain training data, we should rely more on the target domain training data in computing counts related to $w$. So we define a set of words $V_T$ that consists of distinguishable target domain dependent words. A word $w$ belongs to $V_T$ if $P(w|+)$ is much larger or much smaller than $P(w|-)$ in the target domain, i.e., $\frac{{P\left( {w| + } \right)}}{{P\left( {w| - } \right)}} \geq \sigma $ or $\frac{{P\left( {w| - } \right)}}{{P\left( {w| + } \right)}} \geq \sigma $, where $\sigma$ is a parameter. Such words are already effective in classification for the target domain, so the virtual counts in optimization should follow the empirical counts ($N_{+,w}^t$ and $N_{-,w}^t$) in the target domain, which are reflected in the L2 regularization penalty term below ($\alpha $ is the regularization coefficient):

\renewcommand{\arraystretch}{0.9} % Adjust the row spacing.
\renewcommand{\tabcolsep}{16pt} % Adjust the column spacing.
\begin{table*}[t]
	\small
	\begin{center}
		\begin{tabular}{|c|c|c|c|c|c|c|c|}
			\hline
			\rule{0pt}{10pt} NB-T  & NB-S  & NB-ST & SVM-T & SVM-S & SVM-ST & CLF   & \textbf{LSC} \\ \hline
			\rule{0pt}{10pt} 56.21 & 57.04 & 60.61 & 57.82 & 57.64 & 61.05  & 12.87 & \textbf{67.00}  \\
			\hline
		\end{tabular}
	\end{center}
	\vspace{-2mm}
	\caption{\label{table2} Natural class distribution: Average F1-score of the negative class over 20 domains. Negative class is the minority class and thus harder to classify.}
\end{table*}

\renewcommand{\arraystretch}{0.9} % Adjust the row spacing.
\renewcommand{\tabcolsep}{16pt} % Adjust the column spacing.
\begin{table*}[t]
	\small
	\begin{center}
		\begin{tabular}{|c|c|c|c|c|c|c|c|}
			\hline
			\rule{0pt}{10pt} NB-T  & NB-S  & NB-ST & SVM-T & SVM-S & SVM-ST & CLF   & \textbf{LSC}   \\ \hline
			\rule{0pt}{10pt} 80.15 & 77.35 & 80.85 & 78.45 & 78.20  & 79.40   & 80.49 & \textbf{83.34}  \\
			\hline
		\end{tabular}
	\end{center}
	\vspace{-2mm}
	\caption{\label{table3} Balanced class distribution: Average accuracy over 20 domains for each system.}
	\vspace{-3mm}
\end{table*}

\begin{equation} \label{eq:7}
\small
\frac{1}{2}\alpha \mathop \sum \limits_{w \in {V_T}} \left( {{{\left( {{X_{ + ,w}} - N_{ + ,w}^t} \right)}^2} + {{\left( {{X_{ - ,w}} - N_{ - ,w}^t} \right)}^2}} \right)
\end{equation}

To leverage domain-level knowledge (the second type of knowledge in $KB$ in Section~\ref{sec:componentsinlsc}), we want to utilize only those reliable parts of knowledge. The rationale here is that if a word only appears in one or two past domains, the knowledge associated with it is probably not reliable or it is highly specific to those domains. Based on it, we use domain frequency to define the reliability of the domain-level knowledge. For $w$, if $M_{ + ,w}^{KB} \geq \tau $ or $M_{ - ,w}^{KB} \geq \tau $ ($\tau$ is a parameter), we regard it as appearing in a reasonable number of domains, making its knowledge reliable. We denote the set of such words as $V_S$. Then we add the second penalty term as follows:

\begin{equation}
\begin{aligned}
\small
&\frac{1}{2}\alpha \mathop \sum \limits_{w \in {V_S}} {\left( {{X_{ + ,w}} - {R_w} \times X_{ + ,w}^0} \right)^2} \\
& + \frac{1}{2}\alpha \mathop \sum \limits_{w \in {V_S}} {\left( {{X_{ - ,w}} - \left( {1 - {R_w}} \right) \times X_{ - ,w}^0} \right)^2}
\end{aligned}
\end{equation}

\noindent where the ratio $R_w$ is defined as $M_{+,w}^{KB}/(M_{+,w}^{KB}+M_{-,w}^{KB})$. $X_{+,w}^0$ and $X_{-,w}^0$ are the starting points for SGD (Section~\ref{sec:objectivefunction}). Finally, we revise the partial derivatives in Eqs. 4-6 by adding the corresponding partial derivatives of Eqs. 7 and 8 to them. 

\section{Experiments}

\textbf{Datasets}. We created a large corpus containing reviews from 20 types of diverse products or domains crawled from Amazon.com (i.e., 20 datasets). The names of product domains are listed in Table~\ref{tab:20domains}. Each domain contains 1,000 reviews. Following the existing work of other researchers~\cite{Blitzer2007,Pang2002b}, we treat reviews with rating $>$ 3 as positive and reviews with rating $<$ 3 as negative. The datasets are publically available at the authors’ websites. 

\textit{Natural class distribution}: We keep the natural (or skewed) distribution of the positive and negative reviews to experiment with the real-life situation. F1-score is used due to the imbalance.

\textit{Balanced class distribution}: We also created a balance dataset with 200 reviews (100 positive and 100 negative) in each domain dataset. This set is smaller because of the small number of negative reviews in each domain. Accuracy is used for evaluation in this balanced setting.

We used unigram features with no feature selection in classification. We followed~\cite{Pang2002b} to deal with negation words. For evaluation, each domain is treated as the target domain with the rest 19 domains as the past domains. All the models are evaluated using 5-fold cross validation.

\textbf{Baselines}. We compare our proposed LSC model with Na\"{i}ve Bayes (NB), SVM\footnote{http://www.csie.ntu.edu.tw/\~{}cjlin/libsvm/}, and CLF~\cite{li2008multi}. Note that NB and SVM can only work on a single domain data. To have a comprehensive comparison, they are fed with three types of training data:

\begin{enumerate}[leftmargin=*]

\item[a)] labeled training data from the target domain only, denoted by NB-T and SVM-T;

\item[b)] labeled training data from all past source domains only, denoted by NB-S and SVM-S;

\item[c)] merged (labeled) training data from all past domains and the target domain, referred to as NB-ST and SVM-ST.

\end{enumerate}

For LSC, we empirically set $\sigma = 6$ and $\tau = 6$. The learning rate $\lambda$ and regularization coefficient $\alpha$ are set to 0.1 empirically. $\lambda$ is set to 1 for (Laplace) smoothing.

Table~\ref{table2} shows the average F1-scores for the negative class in the natural class distribution, and Table~\ref{table3} shows the average accuracies in the balanced class distribution. We can clearly see that our proposed model LSC achieves the best performance in both cases. In general, NB-S (and SVM-S) are worse than NB-T (and SVM-T), both of which are worse than NB-ST (and SVM-ST). This shows that simply merging both past domains and the target domain data is slightly beneficial. Note that the average F1-score for the positive class is not shown as all classifiers perform very well because the positive class is the majority class (while our model performs slightly better than the baselines). The improvements of the proposed LSC model over all baselines in both cases are statistically significant using paired t-test ($p < 0.01$ compared to NB-ST and CLF, $p < 0.0001$ compared to the others). In the balanced class setting (Table~\ref{table3}), CLF performs better than NB-T and SVM-T, which is consistent with the results in~\cite{li2008multi}. However, it is still worse than our LSC model.

\begin{figure}[t]
	\vspace{0mm}
	\begin{center}
		\centerline{\includegraphics[scale=.35]{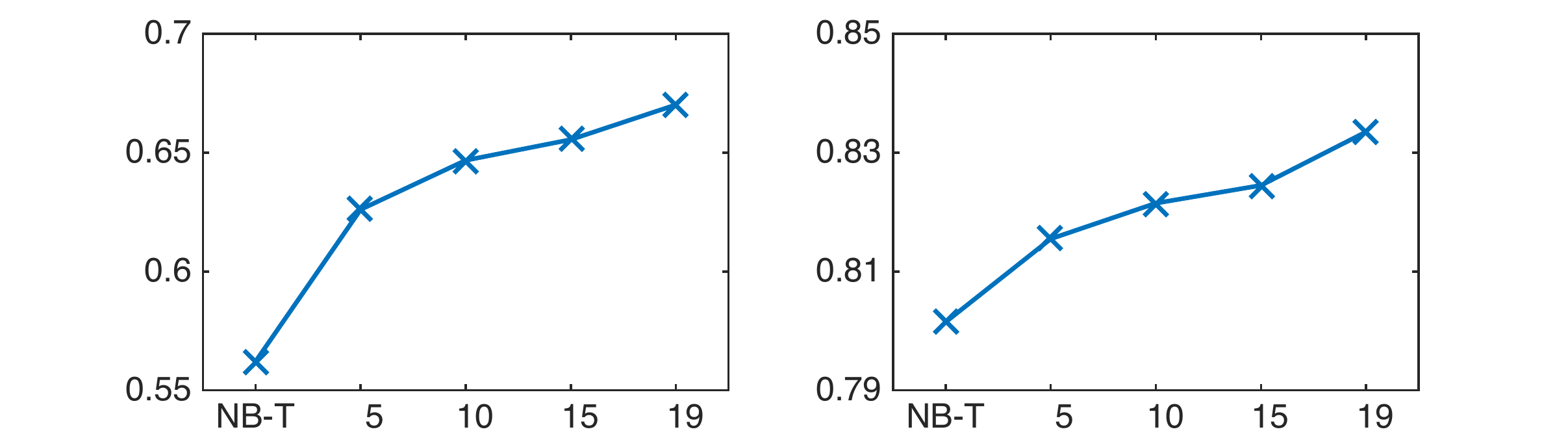}}
		\vspace{-2mm}
		\caption{\label{fig:figure1} (Left): Negative class F1-score of LSC with \#past domains in natural class distribution. (Right): Accuracy of LSC with \#past domains in balanced class distribution.}
		\vspace{-10mm}
	\end{center}
\end{figure}

\textbf{Effects of \#Past Domains}. Figure~\ref{fig:figure1} shows the effects of our model using different number of past domains. We clearly see that LSC performs better with more past domains, showing it indeed has the ability to accumulate knowledge and use the knowledge to build better classifiers.

\section{Conclusions}

In this paper, we proposed a lifelong learning approach to sentiment classification using optimization, which is based on stochastic gradient descent in the framework of Bayesian probabilities. Penalty terms are introduced to effectively exploit the knowledge gained from past learning. Our experimental results using 20 diverse product review domains demonstrate the effectiveness of the method. We believe that lifelong learning is a promising direction for building better classifiers.

\clearpage
\bibliographystyle{acl}
\bibliography{ACL2015Short}

\begin{thebibliography}{}

\bibitem[\protect\citename{Andreevskaia and Bergler}2008]{Andreevskaia2008}
Alina Andreevskaia and Sabine Bergler.
\newblock 2008.
\newblock {When Specialists and Generalists Work Together: Overcoming Domain
  Dependence in Sentiment Tagging.}
\newblock In {\em ACL}, pages 290--298.

\bibitem[\protect\citename{Aue and Gamon}2005]{Aue+Gamon:05a}
Anthony Aue and Michael Gamon.
\newblock 2005.
\newblock {Customizing Sentiment Classifiers to New Domains: A Case Study}.
\newblock In {\em RANLP}.

\bibitem[\protect\citename{Blitzer \bgroup et al.\egroup }2007]{Blitzer2007}
John Blitzer, Mark Dredze, and Fernando Pereira.
\newblock 2007.
\newblock {Biographies, Bollywood, Boom-boxes and Blenders: Domain Adaptation
  for Sentiment Classification.}
\newblock In {\em ACL}, pages 440--447.

\bibitem[\protect\citename{Bollegala \bgroup et al.\egroup
  }2011]{Bollegala2011}
Danushka Bollegala, David~J Weir, and John Carroll.
\newblock 2011.
\newblock {Using Multiple Sources to Construct a Sentiment Sensitive Thesaurus
  for Cross-Domain Sentiment Classification.}
\newblock In {\em ACL HLT}, pages 132--141.

\bibitem[\protect\citename{Bottou}1998]{Bottou1998}
L\'{e}on Bottou.
\newblock 1998.
\newblock Online algorithms and stochastic approximations.
\newblock In David Saad, editor, {\em Online Learning and Neural Networks}.
  Cambridge University Press, Cambridge, UK.
\newblock Oct 2012.

\bibitem[\protect\citename{Carlson \bgroup et al.\egroup }2010]{Carlson2010}
Andrew Carlson, Justin Betteridge, and Bryan Kisiel.
\newblock 2010.
\newblock {Toward an Architecture for Never-Ending Language Learning.}
\newblock In {\em AAAI}, pages 1306--1313.

\bibitem[\protect\citename{Caruana}1997]{Caruana1997}
Rich Caruana.
\newblock 1997.
\newblock {Multitask Learning}.
\newblock {\em Machine learning}, 28(1):41--75.

\bibitem[\protect\citename{Chen and Liu}2014a]{Chen2014KDD}
Zhiyuan Chen and Bing Liu.
\newblock 2014a.
\newblock {Mining Topics in Documents : Standing on the Shoulders of Big Data}.
\newblock In {\em KDD}, pages 1116--1125.

\bibitem[\protect\citename{Chen and Liu}2014b]{Chen2014ICML}
Zhiyuan Chen and Bing Liu.
\newblock 2014b.
\newblock {Topic Modeling using Topics from Many Domains, Lifelong Learning and
  Big Data}.
\newblock In {\em ICML}, pages 703--711.

\bibitem[\protect\citename{Chen \bgroup et al.\egroup }2011]{Chen2011}
Jianhui Chen, Jiayu Zhou, and Jieping Ye.
\newblock 2011.
\newblock {Integrating low-rank and group-sparse structures for robust
  multi-task learning}.
\newblock In {\em KDD}, pages 42--50.

\bibitem[\protect\citename{He \bgroup et al.\egroup }2011]{He2011}
Yulan He, Chenghua Lin, and Harith Alani.
\newblock 2011.
\newblock {Automatically Extracting Polarity-Bearing Topics for Cross-Domain
  Sentiment Classification}.
\newblock In {\em ACL}, pages 123--131.

\bibitem[\protect\citename{Heckman}1979]{heckman1979sample}
James~J Heckman.
\newblock 1979.
\newblock Sample selection bias as a specification error.
\newblock {\em Econometrica: Journal of the econometric society}, pages
  153--161.

\bibitem[\protect\citename{Jiang}2008]{Jiang2008literature}
Jing Jiang.
\newblock 2008.
\newblock {A literature survey on domain adaptation of statistical
  classifiers}.
\newblock Technical report.

\bibitem[\protect\citename{Ku \bgroup et al.\egroup }2009]{Ku2009}
Lun-Wei Ku, Ting-Hao Huang, and Hsin-Hsi Chen.
\newblock 2009.
\newblock {Using morphological and syntactic structures for Chinese opinion
  analysis}.
\newblock In {\em EMNLP}, pages 1260--1269.

\bibitem[\protect\citename{Li and Zong}2008]{li2008multi}
Shoushan Li and Chengqing Zong.
\newblock 2008.
\newblock {Multi-domain sentiment classification}.
\newblock In {\em ACL HLT}, pages 257--260.

\bibitem[\protect\citename{Li \bgroup et al.\egroup }2012]{Li2012}
Fangtao Li, Sinno~Jialin Pan, Ou~Jin, Qiang Yang, and Xiaoyan Zhu.
\newblock 2012.
\newblock {Cross-domain Co-extraction of Sentiment and Topic Lexicons}.
\newblock In {\em ACL}, pages 410--419.

\bibitem[\protect\citename{Li \bgroup et al.\egroup }2013]{Li2013}
Shoushan Li, Yunxia Xue, Zhongqing Wang, and Guodong Zhou.
\newblock 2013.
\newblock {Active learning for cross-domain sentiment classification}.
\newblock In {\em AAAI}, pages 2127--2133.

\bibitem[\protect\citename{Liu}2012]{liu2012}
Bing Liu.
\newblock 2012.
\newblock {Sentiment Analysis and Opinion Mining}.
\newblock {\em Synthesis Lectures on Human Language Technologies}, 5(1):1--167.

\bibitem[\protect\citename{Pan and Yang}2010]{Pan2010TLSurvey}
Sinno~Jialin Pan and Qiang Yang.
\newblock 2010.
\newblock {A Survey on Transfer Learning.}
\newblock {\em IEEE Trans. Knowl. Data Eng.}, 22(10):1345--1359.

\bibitem[\protect\citename{Pang and Lee}2008]{Pang2008Sentiment}
Bo~Pang and Lillian Lee.
\newblock 2008.
\newblock {Opinion mining and sentiment analysis}.
\newblock {\em Foundations and Trends in Information Retrieval}, 2(1-2):1--135.

\bibitem[\protect\citename{Pang \bgroup et al.\egroup }2002]{Pang2002b}
Bo~Pang, Lillian Lee, and Shivakumar Vaithyanathan.
\newblock 2002.
\newblock {Thumbs up? Sentiment Classification using Machine Learning
  Techniques}.
\newblock In {\em EMNLP}, pages 79--86.

\bibitem[\protect\citename{Raina \bgroup et al.\egroup }2007]{Raina2007}
Rajat Raina, Alexis Battle, Honglak Lee, Benjamin Packer, and Andrew~Y Ng.
\newblock 2007.
\newblock {Self-taught Learning : Transfer Learning from Unlabeled Data}.
\newblock In {\em ICML}, pages 759--766.

\bibitem[\protect\citename{Ruvolo and Eaton}2013]{Ruvolo2013ICML}
Paul Ruvolo and Eric Eaton.
\newblock 2013.
\newblock {ELLA: An efficient lifelong learning algorithm}.
\newblock In {\em ICML}, pages 507--515.

\bibitem[\protect\citename{Saha \bgroup et al.\egroup }2011]{saha2011online}
Avishek Saha, Piyush Rai, Suresh Venkatasubramanian, and Hal Daume.
\newblock 2011.
\newblock {Online learning of multiple tasks and their relationships}.
\newblock In {\em AISTATS}, pages 643--651.

\bibitem[\protect\citename{Shimodaira}2000]{shimodaira2000improving}
Hidetoshi Shimodaira.
\newblock 2000.
\newblock {Improving predictive inference under covariate shift by weighting
  the log-likelihood function}.
\newblock {\em Journal of statistical planning and inference}, 90(2):227--244.

\bibitem[\protect\citename{Silver \bgroup et al.\egroup }2013]{Silver2013}
Daniel~L Silver, Qiang Yang, and Lianghao Li.
\newblock 2013.
\newblock {Lifelong Machine Learning Systems: Beyond Learning Algorithms.}
\newblock In {\em AAAI Spring Symposium: Lifelong Machine Learning}, pages
  49--55.

\bibitem[\protect\citename{Tan \bgroup et al.\egroup }2007]{Tan2007}
Songbo Tan, Gaowei Wu, Huifeng Tang, and Xueqi Cheng.
\newblock 2007.
\newblock {A novel scheme for domain-transfer problem in the context of
  sentiment analysis.}
\newblock In {\em CIKM}, pages 979--982.

\bibitem[\protect\citename{Thrun}1998]{Thrun_1998LearningToLearn}
Sebastian Thrun.
\newblock 1998.
\newblock {Lifelong Learning Algorithms}.
\newblock In S~Thrun and L~Pratt, editors, {\em Learning To Learn}, pages
  181--209. Kluwer Academic Publishers.

\bibitem[\protect\citename{Wu \bgroup et al.\egroup
  }2009]{Wu:2009:GRS:1667583.1667681}
Qiong Wu, Songbo Tan, and Xueqi Cheng.
\newblock 2009.
\newblock {Graph Ranking for Sentiment Transfer}.
\newblock In {\em ACL-IJCNLP}, pages 317--320.

\bibitem[\protect\citename{Xia and Zong}2011]{xia2011pos}
Rui Xia and Chengqing Zong.
\newblock 2011.
\newblock {A POS-based Ensemble Model for Cross-domain Sentiment
  Classification.}
\newblock In {\em IJCNLP}, pages 614--622. Citeseer.

\bibitem[\protect\citename{Yoshida \bgroup et al.\egroup
  }2011]{yoshida2011transfer}
Yasuhisa Yoshida, Tsutomu Hirao, Tomoharu Iwata, Masaaki Nagata, and Yuji
  Matsumoto.
\newblock 2011.
\newblock {Transfer Learning for Multiple-Domain Sentiment Analysis-Identifying
  Domain Dependent/Independent Word Polarity}.
\newblock In {\em AAAI}, pages 1286--1291.

\bibitem[\protect\citename{Zadrozny}2004]{zadrozny2004learning}
Bianca Zadrozny.
\newblock 2004.
\newblock {Learning and evaluating classifiers under sample selection bias}.
\newblock In {\em ICML}, page 114. ACM.

\bibitem[\protect\citename{Zhang \bgroup et al.\egroup
  }2008]{zhang2008flexible}
Jian Zhang, Zoubin Ghahramani, and Yiming \textsf{Yang}.
\newblock 2008.
\newblock {Flexible latent variable models for multi-task learning}.
\newblock {\em Machine Learning}, 73(3):221--242.

\end{thebibliography}

\end{document}